\documentclass[10pt,twocolumn,letterpaper]{article}

\usepackage[numbers]{natbib}
\usepackage{cvpr}
\usepackage{times}
\usepackage{epsfig}
\usepackage{graphicx}
\usepackage{amsmath}
\usepackage{amssymb}
\usepackage{multirow}
\usepackage{bm}
\usepackage{booktabs}
\usepackage{subfig}
\usepackage[colorlinks,linkcolor=blue]{hyperref}

\cvprfinalcopy 

\def\cvprPaperID{****} 

\ifcvprfinal\fi
\begin{document}

\title{ccHarmony: Color-checker based Image Harmonization Dataset}

\author{$\textnormal{Haoxu Huang}$, $\textnormal{Li Niu}$\\
 Shanghai Jiao Tong University\\
}

\maketitle
\thispagestyle{empty}

\begin{abstract}
Image harmonization targets at adjusting the foreground in a composite image to make it compatible with the background, producing a more realistic and harmonious image. Training deep image harmonization network requires abundant training data, but it is extremely difficult to acquire training pairs of composite images and ground-truth harmonious images. Therefore, existing works turn to adjust the foreground appearance in a real image to create a synthetic composite image. However, such adjustment may not faithfully reflect the natural illumination change of foreground. In this work, we explore a novel transitive way to construct image harmonization dataset. Specifically, based on the existing datasets with recorded illumination information, we first convert the foreground in a real image to the standard illumination condition, and then convert it to another illumination condition, which is combined with the original background to form a synthetic composite image. In this manner, we construct an image harmonization dataset called ccHarmony, which is named after color checker (cc). The dataset is available at \href{https://github.com/bcmi/Image-Harmonization-Dataset-ccHarmony}{https://github.com/bcmi/Image-Harmonization-Dataset-ccHarmony}.
\end{abstract}


\section{Introduction} \label{sec:intro}

As a prevalent image editing operation, image composition \cite{niu2021making} aims to cut the foreground from one image and paste it on another background image, yielding a composite image. However, the illumination statistics between foreground and background may be inconsistent, making the composite image unrealistic. To solve this issue,
Image harmonization \cite{tsai2017deep,CongDoveNet2020} targets at adjusting the foreground in a composite image to make it compatible with the background, producing a more realistic and harmonious image. Recently, lots of deep learning based image harmonization methods \cite{CongDoveNet2020,Ling2021RegionawareAI,guo2021intrinsic,cong2021high,bargain,sofiiuk2021foreground} have been developed, which greatly advanced the image harmonization performance. 

Training deep learning models require abundant training pairs of composite images and harmonious images. However, it is very tedious and expensive to manually adjust the foreground appearance of composite image to produce harmonious image. Moreover, the reliability of manually created harmonious images is also questionable. Therefore, recent works \cite{tsai2017deep,CongDoveNet2020} resort to an inverse approach, that is, adjusting the foreground in a real image to create a synthetic composite image. In this way, a large number of pairs of synthetic composite images and ground-truth real images can be achieved. However, the synthetic composite images created in this way may have unrealistic foreground with notable artifacts or unreasonable albedo change (\emph{e.g.}, a red vase is converted to a green vase), which would adversely affect the performance of trained harmonization model. Thus, when constructing the image harmonization dataset iHarmony4 \cite{CongDoveNet2020}, the authors adopted both automatic filtering and manual filtering to filter out those unqualified synthetic composite images. Even though, we argue that the obtained synthetic composite images may not faithfully reflect the illumination variation of foreground, leading to non-negligible gap between synthetic composite images and real composite images.

\begin{figure}[t]
  \centering
  \includegraphics[width=\linewidth]{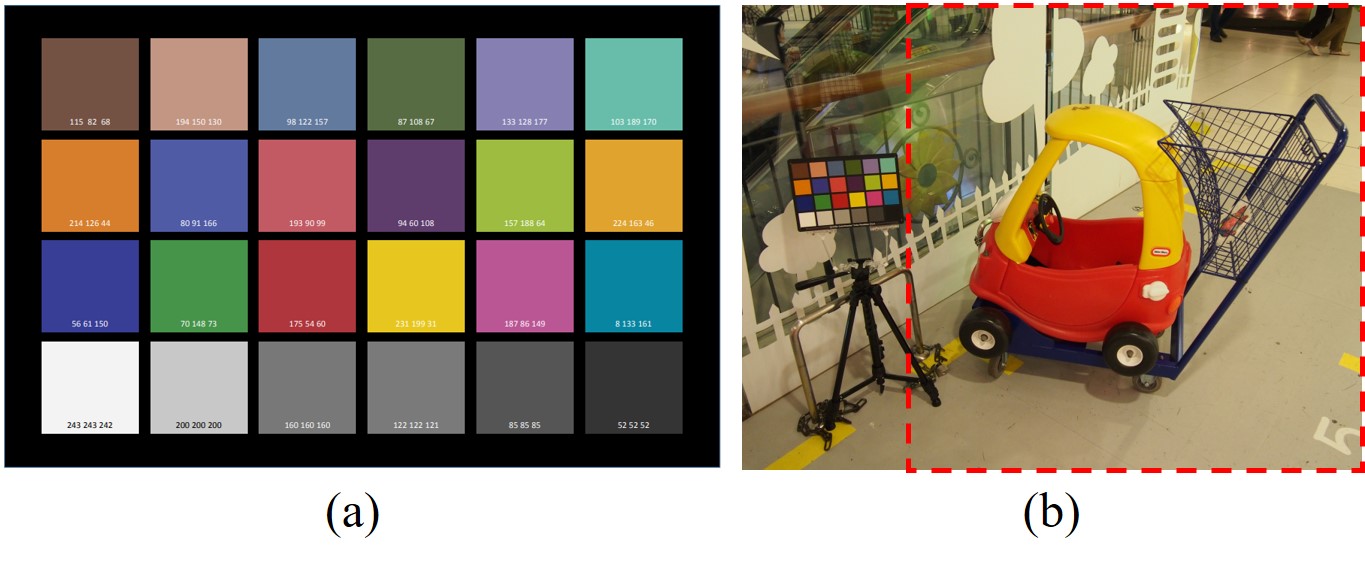}
   \caption{(a) Standard patch colors of a 24-patch Macbeth color checker. (b) An image captured with a color checker placed in the scene. The red dashed box indicates the cropped image without color checker, which is used in our dataset.}
   \label{fig:colorchecker}
\end{figure}

The ideal way to obtain composite images is demonstrated by the Hday2night subdataset in iHarmony4 \cite{CongDoveNet2020}, which is built upon the transient attributes database \cite{laffont2014transient}. The transient attributes database consists of thousands of images from 101 webcams, in which each webcam records the variations within a scene over time. For each scene, we can have a group of images captured in different illumination conditions (\emph{e.g.}, weather, time-of-the-day). Then, the foregrounds are exchanged between two images from the same group, leading to pairs of composite images and real images. However, it is rather difficult to capture exactly the same scene in different illumination conditions. To the best of our knowledge, there are few such datasets like \cite{laffont2014transient} and the number of scenes is very limited.

\begin{figure}[t]
  \centering
  \includegraphics[width=\linewidth]{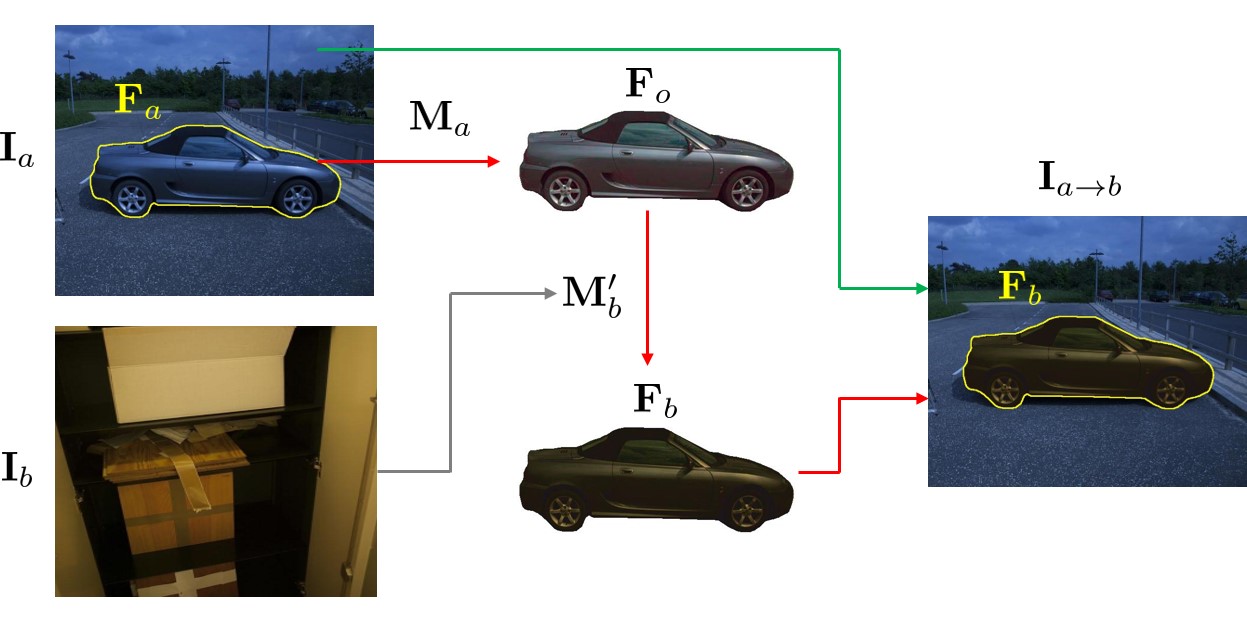}
   \caption{Given a real image $\mathbf{I}_a$, we convert its foreground $\mathbf{F}_a$ to $\mathbf{F}_o$ in standard illumination condition using polynomial matching matrix $\mathbf{M}_a$. Then, we convert $\mathbf{F}_o$ to $\mathbf{F}_b$ in the illumination condition of reference image $\mathbf{I}_b$ using the inverse polynomial matching matrix $\mathbf{M}'_b$. Finally, $\mathbf{F}_b$ is combined with the background of $\mathbf{I}_a$ to produce a synthetic composite image $\mathbf{I}_{a\rightarrow b}$.}
   \label{fig:construction_flowchart}
\end{figure}

In this work, we explore a novel transitive way to construct image harmonization dataset, aiming to simulate the natural illumination variation. Specifically, based on the existing datasets with recorded illumination information, we first convert the foreground in a real image to the standard illumination condition, and then convert it to another illumination condition, which is combined with the original background to produce a synthetic composite image. For the datasets with recorded illumination information, we find two publicly available illumination estimation datasets: NUS dataset \cite{cheng2014illuminant} and Gehler dataset \cite{gehler2008bayesian}. In these two datasets, images are captured using DSLR cameras with a color checker placed in the scene that provides ground truth reference for illumination estimation. Given a 24-patch Macbeth color checker, we have the original colors of 24 patches in standard illumination condition (see Figure \ref{fig:colorchecker}(a)), which is referred to as standard patch color in this paper. Given an image with color checker, we can automatically extract the colors of 24 patches after localizing the color checker, which is referred to as image patch color in this paper. Inspired by previous works \cite{zhou2018color,luo2021estimating,afifi2019color} on white balance or color constancy, we use polynomial matching to characterize the color transformation between standard patch colors and image patch colors. Formally, we denote the standard patch colors of 24 patches as $\mathcal{C}_o=\{\mathbf{c}^o_1,\mathbf{c}^o_2,\ldots,\mathbf{c}^o_{24}\}$ and image patch colors of 24 patches in image $\mathbf{I}_a$ as  $\mathcal{C}_a=\{\mathbf{c}^a_1,\mathbf{c}^a_2,\ldots,\mathbf{c}^a_{24}\}$. We can learn a polynomial matching matrix $\mathbf{M}_a$ which transforms from $\mathcal{C}_a$ to $\mathcal{C}_o$. We can also learn an inverse polynomial matching matrix $\mathbf{M}'_a$ which transforms from $\mathcal{C}_o$ to $\mathcal{C}_a$. Intuitively, a polynomial matching matrix represents one illumination condition. Provided with abundant images, we can get abundant polynomial matching matrices representing diverse illumination conditions. One the premise of these polynomial matching matrices, we can transfer across different illumination conditions to simulate natural illumination variation. 

After introducing the background knowledge of color checker and polynomial matching, we briefly describe our way of creating synthetic composite images. As depicted in Figure~\ref{fig:construction_flowchart}, given an image $\mathbf{I}_a$ with foreground mask, we convert its foreground $\mathbf{F}_a$ to the one $\mathbf{F}_o$ in standard illumination condition using polynomial matching matrix $\mathbf{M}_a$. Next, by using the inverse polynomial matching matrix $\mathbf{M}'_b$ of another reference image $\mathbf{I}_b$, we convert $\mathbf{F}_o$ to $\mathbf{F}_b$ in the illumination condition of $\mathbf{I}_b$. Finally, we replace the foreground  $\mathbf{F}_a$ in $\mathbf{I}_a$ with its counterpart $\mathbf{F}_b$, yielding a synthetic composite image $\mathbf{I}_{a\rightarrow b}$. In this manner, we can acquire adequate pairs of synthetic composite images and real images like $\{\mathbf{I}_{a\rightarrow b}, \mathbf{I}_a\}$. Because our dataset is constructed based on images with color checker (cc), we name our dataset as ccHarmony. In the next section, we will describe the details of dataset construction process.

\begin{figure*}[t]
  \centering
  \includegraphics[width=0.95\linewidth]{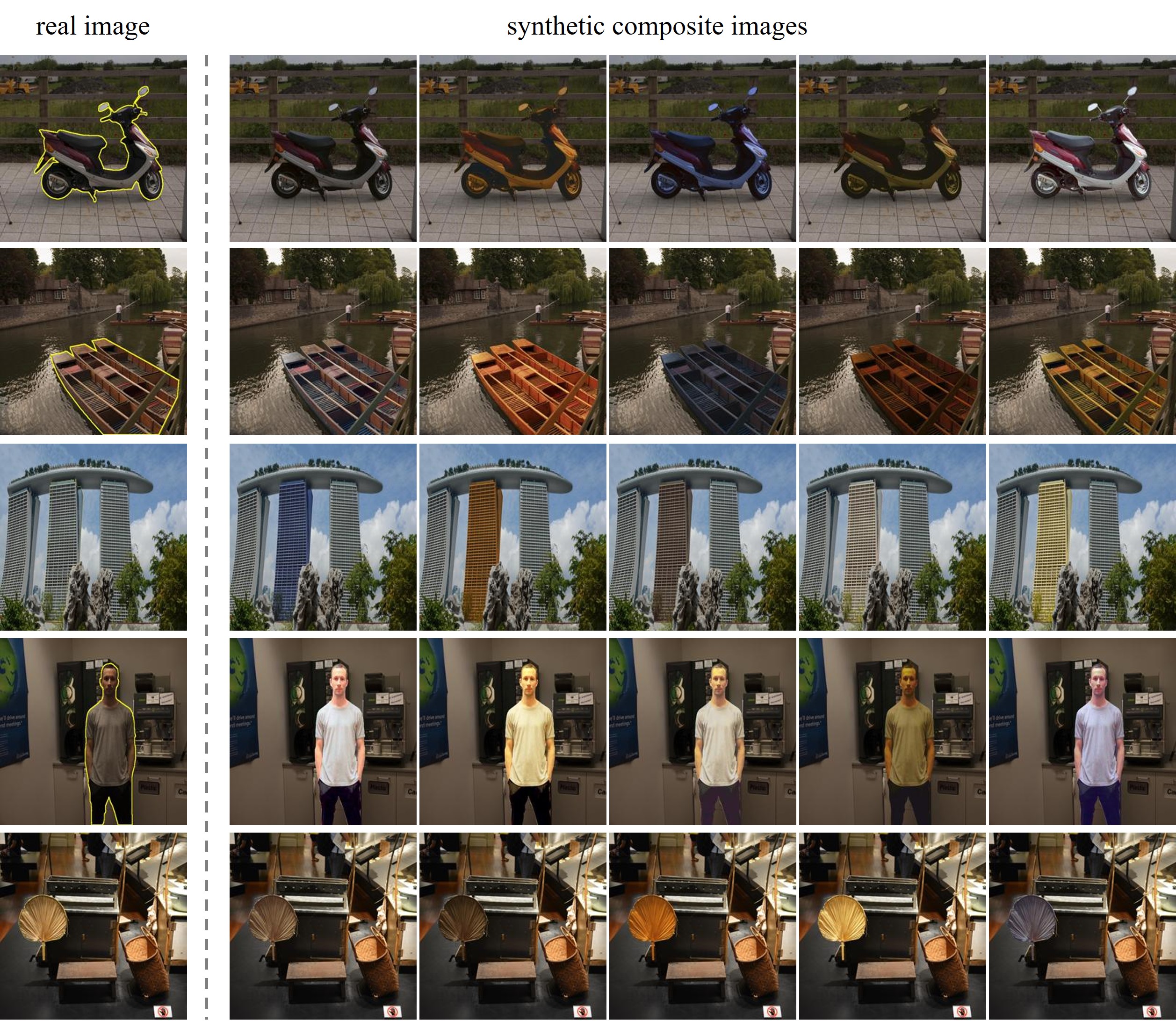}
   \caption{The leftmost column shows the real images in our ccHarmony dataset, in which the foregrounds are outlined in yellow. The rest columns show their corresponding synthetic composite images. }
   \label{fig:example_images}
\end{figure*}

\section{Dataset Construction}

When constructing our ccHarmony dataset, we collect real images with color checker, segment proper foregrounds, and perform color transfer for the foregrounds, yielding synthetic composite images.

\subsection{Real Image Selection}

We first collect images with color checker from NUS
dataset \cite{cheng2014illuminant} and Gehler dataset \cite{gehler2008bayesian}, in which images are captured with a color checker placed in the scene to record illumination information. Then, we perform the following filtering steps.
1) We notice that these two datasets contain images capturing the same scene with similar camera viewpoints, so we perform near-duplicate removal to remove the images with duplicated content. 2) We observe that in some images, the color checker cannot represent the global illumination information of the whole image, for example, the color checker is placed in the shadow area. Therefore, we remove those images with misleading color checker. 3) Another issue is that the color checker should not be included in the final image harmonization dataset, because the color check may provide shortcut for the harmonization network. Therefore, we discard the images in which the color checker is placed near the image center, and crop the remaining images to obtain the possibly largest region without color checker (see Figure \ref{fig:colorchecker}(b)). After the above filtering steps, we have $350$ real images.
 
\subsection{Foreground Segmentation}
 
For each real image, we manually segment one or two foregrounds. 
When selecting foregrounds, we ensure that the color checker can roughly represent the illumination information of the foreground, so that it is meaningful to apply the polynomial matching matrix calculated based on the color checker to the foreground. In total, we segment $426$ foregrounds in $350$ real images, in which the foregrounds cover a wide range of categories like human, tree, building, furniture, staple goods, and so on (see Figure \ref{fig:example_images}).

\subsection{Foreground Color Transfer}

As described in Section~\ref{sec:intro}, given an image $\mathbf{I}_a$, we first  calculate polynomial matching matrix $\mathbf{M}_a$ according to its color checker. Then, we apply $\mathbf{M}_a$ to the foreground $\mathbf{F}_a$ in $\mathbf{I}_a$ to convert it to $\mathbf{F}_o$, which is expected to be its counterpart in standard illumination condition. 

Next, we randomly select $10$ other real images as reference images for $\mathbf{I}_a$. For each reference image $\mathbf{I}_b$, we first calculate inverse polynomial matching matrix $\mathbf{M}'_b$ according to its color checker. Then, we apply $\mathbf{M}'_b$ to $\mathbf{F}_o$ to convert it to $\mathbf{F}_b$, which is expected to be its counterpart in the illumination condition of $\mathbf{I}_b$. Finally, we combine $\mathbf{F}_b$ and the background of $\mathbf{I}_a$ to form a synthetic composite image $\mathbf{I}_{a\rightarrow b}$. The above procedure is illustrated in Figure~\ref{fig:construction_flowchart}.
 Since we select $10$ reference images for each foreground, we can produce $10$ synthetic composite images for each foreground. Based on $426$ foregrounds, we can produce $4260$ pairs of synthetic composite images and real images. 

We split $350$ images into $250$ training images with $308$ foregrounds and $10$ test images with $118$ foregrounds. Thus, the training set contains $3080$ pairs of synthetic composite images and real images, while the test set contains $1180$ pairs. We show several examples of real images and their corresponding synthetic composite images in Figure \ref{fig:example_images}. 
 
\section{Conclusion}
In this paper, we explore a novel transitive way to construct image harmonization dataset based on the images with recorded illumination information. Specifically, we convert the foreground in a real image to the standard illumination condition, and then convert it to another illumination condition, leading to a synthetic composite image. We have released our ccHarmony dataset and hope that this dataset can facilitate the research in the field of image harmonization.

{\small
\bibliographystyle{ieee_fullname}
\bibliography{main.bbl}
}

\end{document}